\documentclass[conference]{IEEEtran}
\usepackage{booktabs}
\usepackage{cite}
\usepackage{amsmath,amssymb,amsfonts}
\usepackage{algorithmic}
\usepackage{graphicx}
\usepackage{textcomp}
\usepackage{xcolor}
\def\BibTeX{{\rm B\kern-.05em{\sc i\kern-.025em b}\kern-.08em
    T\kern-.1667em\lower.7ex\hbox{E}\kern-.125emX}}
    
\IEEEoverridecommandlockouts  % 确保 IEEEoverridecommandlockouts 位置正确

\begin{document}

\title{Performance Analysis of Traditional VQA Models Under Limited Computational Resources}

\author{
    \IEEEauthorblockN{Jihao Gu}
    \IEEEauthorblockA{
        \textit{Department of Computer Science} \\
        \textit{University College London}\\
        London, UK \\
        jihao.gu.23@ucl.ac.uk
    }
}

\maketitle

\begin{abstract}
In real-world applications where computational resources are limited, effectively integrating visual and textual information for Visual Question Answering (VQA) presents significant challenges. This paper investigates the performance of traditional models under computational constraints, focusing on enhancing VQA performance, particularly for numerical and counting questions. We evaluate models based on Bidirectional GRU (BidGRU), GRU, Bidirectional LSTM (BidLSTM), and Convolutional Neural Networks (CNN), analyzing the impact of different vocabulary sizes, fine-tuning strategies, and embedding dimensions. Experimental results show that the BidGRU model with an embedding dimension of 300 and a vocabulary size of 3000 achieves the best overall performance without the computational overhead of larger models. Ablation studies emphasize the importance of attention mechanisms and counting information in handling complex reasoning tasks under resource limitations. Our research provides valuable insights for the development of more efficient VQA models suitable for deployment in environments with limited computational capacity.
\end{abstract}


% ===============================================================================

\section{Introduction}
In real-world applications, computational resources are often limited, especially in fields like medicine and industrial automation, where deploying large-scale deep learning models is impractical. Under such constraints, traditional models still play a crucial role in tasks that require both efficiency and accuracy. Visual Question Answering (VQA) is one such task that demands the integration of visual and textual information to answer questions based on images, posing significant challenges when resources are constrained.

This study investigates the performance of traditional models in addressing VQA tasks under limited computational resources. We focus on understanding how various question feature extraction methods and model configurations impact efficiency and accuracy, particularly for numerical and counting questions. By exploring models based on Bidirectional GRU (BidGRU), GRU, Bidirectional LSTM (BidLSTM), and Convolutional Neural Networks (CNN), we aim to identify strategies that optimize performance without the overhead of large-scale models.

Our contributions include a comprehensive analysis of traditional models' adaptability in constrained environments and actionable insights for practitioners working in resource-limited scenarios. Through detailed experimental analysis, we demonstrate that certain configurations, such as the BidGRU model with specific embedding dimensions and vocabulary sizes, can achieve superior performance. These findings provide valuable guidance for developing more efficient VQA models suitable for deployment in environments with limited computational capacity.

% ===============================================================================

\section{Related Work}

% -------------------------------------------------------------------------------
\subsection{Visual Question Answering}
Visual Question Answering (VQA) is an interdisciplinary task that combines computer vision and natural language processing. It requires models to answer questions based on visual content. Various approaches have been proposed to improve VQA performance through advanced attention mechanisms and neural architectures.

\begin{itemize}
    \item \textbf{Spatial Memory Network}: Employs a two-hop attention mechanism. The first hop aligns question words with image regions, capturing detailed local evidence. The second hop refines this evidence by considering the entire question embedding, enhancing prediction accuracy \cite{chen2018}.
    \item \textbf{BIDAF Model}: Utilizes a bi-directional attention mechanism to create query-aware context representations, capturing interactions between context and query \cite{seo2017}.
    \item \textbf{CNN for Text Representation}: Replaces RNNs with CNNs for text representation in VQA, demonstrating superior capability in capturing textual features \cite{noh2016}.
    \item \textbf{Structured Attentions}: Models visual attention as a multivariate distribution over a conditional random field (CRF) to better encode relationships between multiple image regions \cite{zhang2018}.
    \item \textbf{Inverse VQA (iVQA)}: Introduces the inverse VQA task, using question-ranking-based evaluation to diagnose model strengths and weaknesses \cite{gordon2018}.
\end{itemize}
% -------------------------------------------------------------------------------
\begin{figure*}[!t]
    \centering
    \includegraphics[width=1.0\textwidth]{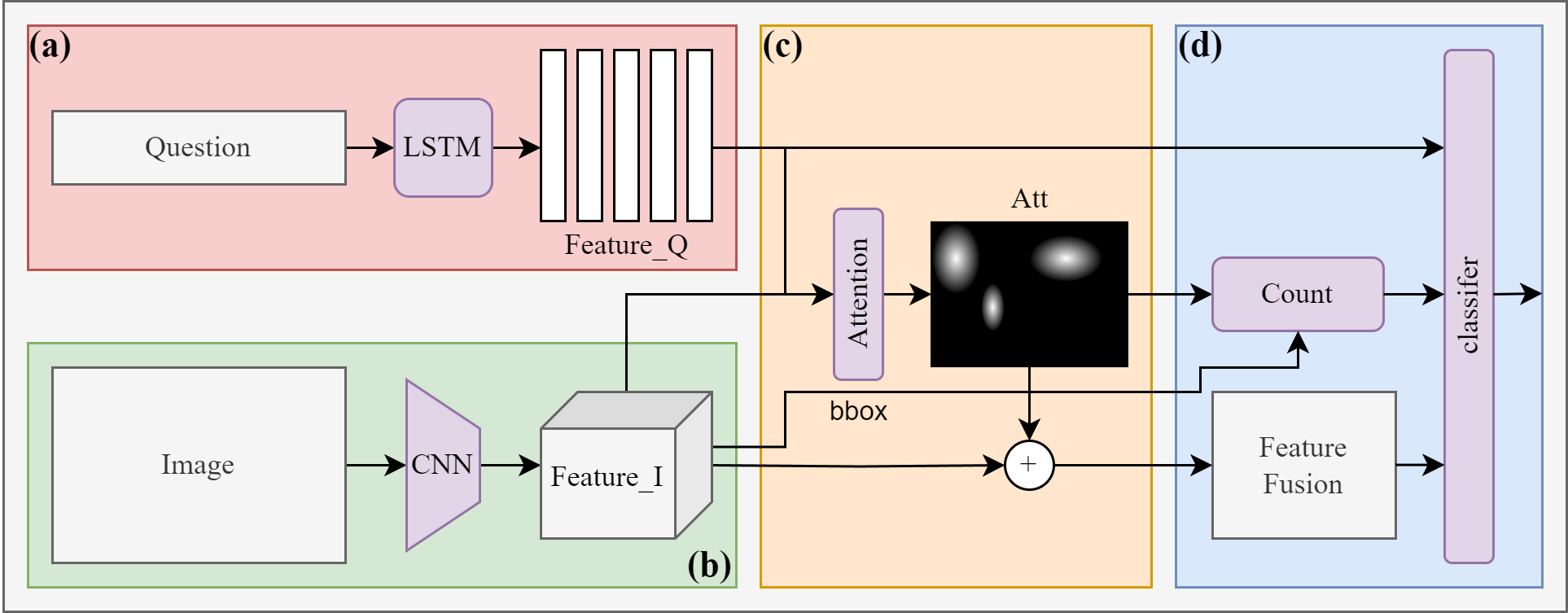} % 请确保图片文件名和路径正确
    \caption{The VQA model architecture consisting of (a) question feature extraction, (b) image feature extraction, (c) attention mechanism, and (d) feature fusion and classification modules.}
    \label{fig:framework}
\end{figure*}
% -------------------------------------------------------------------------------
\subsection{Image Captioning}
Image captioning generates descriptive textual information for images, requiring models to understand and describe visual content accurately.

\begin{itemize}
    \item \textbf{Show, Attend and Tell}: Integrates CNNs with LSTMs and uses attention mechanisms to focus on relevant image regions for accurate captions \cite{xu2015}.
    \item \textbf{Self-Critical Sequence Training (SCST)}: Uses the REINFORCE algorithm for reinforcement learning, optimizing the CIDEr metric to reduce exposure bias \cite{rennie2017}.
    \item \textbf{Meshed-Memory Transformer (M2)}: Employs multi-level encoding and memory-augmented attention for improved caption generation \cite{cornia2020}.
    \item \textbf{X-Linear Attention Networks (X-LAN)}: Captures second-order interactions using bilinear pooling, enhancing feature representation \cite{pan2020}.
\end{itemize}
% -------------------------------------------------------------------------------
\subsection{Multi-Modal}
Multimodal research focuses on models that integrate information from multiple modalities, such as text, images, and audio, to perform complex tasks.

\begin{itemize}
    \item \textbf{LayoutLMv2}: Integrates text, layout, and image information through a multi-modal Transformer architecture, enhancing document understanding \cite{xu2020}.
    \item \textbf{Cross-Modal Context for Image Captioning}: Combines textual and visual contextual information using CLIP and Visual Genome datasets \cite{mokady2021}.
    \item \textbf{Transformer-Based Multi-Modal Proposal and Re-Rank}: Uses CLIP and XLM-RoBERTa for image-caption matching through a multi-modal approach \cite{li2021}.
\end{itemize}

% ===============================================================================

\section{Method}

In this section, we detail the architecture of our Visual Question Answering (VQA) model, which is designed to effectively integrate visual and textual information for accurate answer prediction. The model consists of four main components: (a) question feature extraction module, (b) image feature extraction module, (c) attention mechanism module, and (d) feature fusion and classification module. ~\ref{fig:framework} illustrates the overall structure of the model.
% -------------------------------------------------------------------------------
\subsection{Overview}
Our model takes an image $V$ and a question $Q$ as inputs and predicts the correct answer $A$. The process involves:

\begin{itemize}
    \item \textbf{Question Feature Extraction}: Encoding the question using a BidGRU.
    \item \textbf{Image Feature Extraction}: Processing the image with pre-trained CNNs.
    \item \textbf{Attention Mechanism}: Highlighting relevant image regions based on the question.
    \item \textbf{Counting Module}: Estimating the number of relevant objects for counting questions.
    \item \textbf{Feature Fusion and Classification}: Combining visual, textual, and counting features to predict the answer.
\end{itemize}
% -------------------------------------------------------------------------------
\begin{table*}[!t]
\centering
\caption{Performance Comparison of Different Question Feature Extraction Methods}
\label{tab:1}
\resizebox{\textwidth}{!}{
\begin{tabular}{lccccccc}
\toprule
\textbf{Model} & \textbf{number (s)} & \textbf{number (p)} & \textbf{count (s)} & \textbf{count (p)} & \textbf{Other} & \textbf{all (s)} & \textbf{all (p)} \\
\midrule
\textbf{BidGRU (Dim=300)} & \textbf{49.52\%} & \textbf{23.25\%} & \textbf{57.20\%} & \textbf{26.80\%} & \textbf{57.33\%} & \textbf{65.74\%} & \textbf{37.83\%} \\
GRU (From Paper) & 49.59\% & 23.37\% & 57.29\% & 26.96\% & 57.18\% & 65.42\% & 37.25\% \\
BidGRU (2048-1024) & 49.39\% & 23.09\% & 57.13\% & 26.69\% & 56.88\% & 65.37\% & 37.27\% \\
BidLSTM (Dim=300) & 47.39\% & 20.40\% & 54.73\% & 23.56\% & 56.78\% & 64.92\% & 36.67\% \\
GRU (300 MFB k=5) & 46.27\% & 19.30\% & 53.57\% & 22.29\% & 55.98\% & 64.13\% & 35.96\% \\
CNN (kernel=3+4+5) & 41.08\% & 13.55\% & 47.32\% & 15.62\% & 54.41\% & 61.86\% & 32.49\% \\
\bottomrule
\end{tabular}
}
\end{table*}
% -------------------------------------------------------------------------------
\subsection{Question Feature Extraction}
The question feature extraction module aims to convert the variable-length question \( Q \) into a fixed-dimensional feature vector that encapsulates its semantic meaning.

\subsubsection{Embedding Layer}
We use an embedding layer to convert each word in the question into a continuous vector representation:
\begin{equation}
\mathbf{E} = [\mathbf{e}_1, \mathbf{e}_2, \dots, \mathbf{e}_T], \quad \mathbf{e}_t \in \mathbb{R}^d
\label{eq:embedding}
\end{equation}
where \( T \) is the maximum question length, and \( d \) is the embedding dimension (set to 300 in our experiments). The embedding layer is initialized randomly and learned during training.

\subsubsection{Bidirectional GRU Encoder}
To capture contextual information, we employ a Bidirectional GRU:
\begin{equation}
\overrightarrow{\mathbf{h}}_t = \text{GRU}(\mathbf{e}_t, \overrightarrow{\mathbf{h}}_{t-1}), \quad 
\overleftarrow{\mathbf{h}}_t = \text{GRU}(\mathbf{e}_t, \overleftarrow{\mathbf{h}}_{t+1})
\label{eq:gru_forward_backward}
\end{equation}

The final question representation \( \mathbf{q} \) is obtained by concatenating the last hidden states from both directions:
\begin{equation}
\mathbf{q} = [\overrightarrow{\mathbf{h}}_T; \overleftarrow{\mathbf{h}}_1] \in \mathbb{R}^{2h}
\label{eq:final_question_representation}
\end{equation}
where \( h \) is the hidden size of the GRU.
% -------------------------------------------------------------------------------
\subsection{Image Feature Extraction}
For image feature extraction, we use pre-trained convolutional neural networks (CNNs) to obtain spatial feature maps from the input image:
\begin{equation}
\mathbf{V} = \text{CNN}(V) \in \mathbb{R}^{C \times H \times W}
\label{eq:image_features}
\end{equation}
where \( C \) is the number of feature channels, and \( H \times W \) is the spatial dimension.

\subsubsection{Feature Normalization}
The visual features are normalized to unit length to ensure numerical stability:
\begin{equation}
\mathbf{V} = \frac{\mathbf{V}}{\|\mathbf{V}\|_2 + \epsilon}
\label{eq:feature_normalization}
\end{equation}
where \( \epsilon \) is a small constant to avoid division by zero.
% -------------------------------------------------------------------------------
\subsection{Attention Mechanism}
The attention mechanism aligns the question representation with relevant image regions, allowing the model to focus on pertinent visual information.

\subsubsection{Attention Computation}
We compute attention maps using the following steps:
\begin{itemize}
    \item \textbf{Linear Transformations}: Project the visual and question features into a common space:
    \begin{equation}
    \mathbf{V}' = \text{Conv2d}(\mathbf{V}), \quad \mathbf{q}' = \text{Linear}(\mathbf{q})
    \label{eq:linear_projection}
    \end{equation}
    
    \item \textbf{Feature Tiling}: Tile the question features over the spatial dimensions of the visual features:
    \begin{equation}
    \mathbf{Q} = \text{Tile}(\mathbf{q}', H, W)
    \label{eq:feature_tiling}
    \end{equation}
    
    \item \textbf{Fusion}: Combine the visual and question features using a fusion function \( f \):
    \begin{equation}
    \mathbf{F} = f(\mathbf{V}', \mathbf{Q})
    \label{eq:fusion_function}
    \end{equation}
    We employ a custom fusion function:
    \begin{equation}
    f(\mathbf{x}, \mathbf{y}) = -(\mathbf{x} - \mathbf{y})^2 + \text{ReLU}(\mathbf{x} + \mathbf{y})
    \label{eq:custom_fusion}
    \end{equation}
    
    \item \textbf{Attention Map Generation}: Apply a convolution to produce the attention maps:
    \begin{equation}
    \mathbf{A} = \text{Conv2d}(\mathbf{F}) \in \mathbb{R}^{G \times H \times W}
    \label{eq:attention_map}
    \end{equation}
    where \( G \) is the number of glimpses (set to 2).
\end{itemize}

\subsubsection{Attention Application}
We apply the attention maps to the visual features to obtain attended features:
\begin{equation}
\mathbf{V}_{\text{att}} = \sum_{i=1}^{G} \text{Softmax}(\mathbf{A}_i) \odot \mathbf{V}
\label{eq:attended_features}
\end{equation}
where \( \odot \) denotes element-wise multiplication.
% -------------------------------------------------------------------------------
\subsection{Counting Module}
To enhance the model's ability to handle counting questions, we incorporate a counting module that estimates the number of objects in the image relevant to the question.
% -------------------------------------------------------------------------------
\begin{table*}[!t]
\centering
\caption{Performance Comparison with Different Token Sizes}
\label{tab:2}
\resizebox{\textwidth}{!}{
\begin{tabular}{lcccccccc}
\toprule
\textbf{Model} & \textbf{Token Size} & \textbf{number (s)} & \textbf{number (p)} & \textbf{count (s)} & \textbf{count (p)} & \textbf{Other} & \textbf{all (s)} & \textbf{all (p)} \\
\midrule
BidGRU (Dim=300) & 2000 & 49.23\% & 22.70\% & 56.87\% & 26.18\% & 56.96\% & 65.46\% & 37.54\% \\
\textbf{BidGRU (Dim=300)} & \textbf{3000} & \textbf{49.52\%} & \textbf{23.25\%} & \textbf{57.20\%} & \textbf{26.80\%} & \textbf{57.33\%} & \textbf{65.74\%} & \textbf{37.83\%} \\
BidGRU (Dim=300) & 4000 & 49.16\% & 22.95\% & 56.83\% & 26.51\% & 57.33\% & 65.65\% & 37.61\% \\
\bottomrule
\end{tabular}
}
\end{table*}
% -------------------------------------------------------------------------------
\begin{table*}[!t]
\centering
\caption{Performance Comparison of Different Model Fine-Tuning Strategies}
\label{tab:3}
\resizebox{\textwidth}{!}{
\begin{tabular}{lcccccccc}
\toprule
\textbf{Model} & \textbf{Fine-Tuning Strategy} & \textbf{number (s)} & \textbf{number (p)} & \textbf{count (s)} & \textbf{count (p)} & \textbf{Other} & \textbf{all (s)} & \textbf{all (p)} \\
\midrule
BidGRU (Dim=300) & Baseline & \textbf{49.52\%} & \textbf{23.25\%} & \textbf{57.20\%} & \textbf{26.80\%} & \textbf{57.33\%} & \textbf{65.74\%} & \textbf{37.83\%} \\
+ Dropout 0.5 & Dropout & 48.83\% & 22.36\% & 56.47\% & 25.85\% & 57.18\% & 65.51\% & 37.41\% \\
Hidden (2048-1024) & Adjusted Dimensions & 49.39\% & 23.09\% & 57.13\% & 26.69\% & 56.88\% & 65.37\% & 37.27\% \\
+ Soft Labels & Soft Labels & 49.24\% & 22.92\% & 56.92\% & 26.36\% & 57.05\% & 65.25\% & 37.06\% \\
\bottomrule
\end{tabular}
}
\end{table*}
% -------------------------------------------------------------------------------
\subsubsection{Counting Features}
We use the first attention map \( \mathbf{A}_1 \) and bounding box information \( \mathbf{b} \) to compute counting features:
\begin{equation}
\mathbf{c} = \text{Counter}(\mathbf{b}, \mathbf{A}_1)
\label{eq:counting_features}
\end{equation}
The counting module processes the attention weights and spatial information to produce a count feature vector.
% -------------------------------------------------------------------------------
\subsection{Feature Fusion and Classification}
The final step combines the attended visual features, question features, and counting information to predict the answer.

\subsubsection{Fusion Mechanism}
We fuse the features using a combination of linear transformations and the fusion function:
\begin{equation}
\mathbf{x} = f(\text{Linear}(\mathbf{V}_{\text{att}}), \text{Linear}(\mathbf{q}))
\label{eq:fusion_combination}
\end{equation}
We also incorporate the counting features:
\begin{equation}
\mathbf{x} = \mathbf{x} + \text{BatchNorm}(\text{ReLU}(\text{Linear}(\mathbf{c})))
\label{eq:counting_integration}
\end{equation}

\subsubsection{Answer Prediction}
The fused features are passed through a final linear layer to produce the answer probabilities:
\begin{equation}
\mathbf{A} = \text{Softmax}(\text{Linear}(\mathbf{x}))
\label{eq:answer_prediction}
\end{equation}
The model is trained using a cross-entropy loss function.

\subsection{Summary}
Our proposed model effectively integrates visual and textual information through a carefully designed architecture that leverages Bidirectional GRUs for question encoding, attention mechanisms for focusing on relevant image regions, and a counting module for handling counting questions. The fusion of features enables the model to perform complex reasoning required for accurate Visual Question Answering.

% ===============================================================================

\section{Experimental Results and Analysis}

In this section, we present a comprehensive analysis and discussion of the experimental results on the Visual Question Answering (VQA) task. We investigate the effects of different model architectures, vocabulary sizes (token sizes), question feature extraction methods, and model fine-tuning strategies on performance. Besides the overall accuracy (\textit{All}), we pay special attention to fine-grained metrics such as \textit{number (single)}, \textit{number (pair)}, \textit{count (single)}, \textit{count (pair)}, \textit{all (single)}, and \textit{all (pair)}. These metrics help us understand the models' capabilities in handling different types and complexities of questions.
% -------------------------------------------------------------------------------
\subsection{Comparison of Question Feature Extraction Methods}
We first compare the impact of different question feature extraction methods on model performance, including models based on Bidirectional GRU (BidGRU), GRU, Bidirectional LSTM (BidLSTM), and CNN. The experimental results are shown in Table~\ref{tab:1}.

\textbf{Analysis}: The \textbf{BidGRU Embedding Dim 300} model achieved the best performance across all metrics, especially in fine-grained indicators like \texttt{number (single)}, \texttt{number (pair)}, \texttt{count (single)}, and \texttt{count (pair)}. This indicates that BidGRU effectively captures the contextual information of questions and has strong understanding and reasoning capabilities for numerical and counting questions. The \textbf{GRU (Reproduced from Paper)} model performed slightly lower than the BidGRU model but had a slight advantage in the \texttt{number (single)} metric. This might be because the unidirectional GRU has certain benefits when handling simple numerical questions. The \textbf{BidLSTM Embedding Dim 300} and \textbf{GRU 300 MFB k=5} models showed lower performance across all metrics compared to the BidGRU model, especially in paired questions (\texttt{number (pair)}, \texttt{count (pair)}), indicating that these models are less capable in handling complex comparison and reasoning questions. The \textbf{CNN kernel=3+4+5} model had the lowest overall performance, with significantly lower accuracy in fine-grained metrics, particularly in paired questions. This suggests that CNNs have limitations in capturing sequential information and dealing with complex language structures in VQA tasks.

\textbf{Conclusion}: The BidGRU model, due to its bidirectional structure, can better understand the semantics and contextual information of questions, performing better in handling numerical and counting questions, particularly in complex comparison and reasoning scenarios involving paired questions.
% -------------------------------------------------------------------------------
\subsection{Impact of Token Size}
We investigated the effect of different vocabulary sizes (token sizes) on model performance. The experimental results are presented in Table~\ref{tab:2}.

\textbf{Analysis}: With a \textbf{token size of 3000}, the model achieved the best performance across all metrics, suggesting that an appropriate vocabulary size helps the model fully learn textual features. A \textbf{token size of 2000} led to decreased performance, possibly due to the vocabulary being too small, resulting in some key words being excluded and limiting the model's expressive capacity. Increasing the \textbf{token size to 4000} did not further improve performance; some metrics slightly decreased, possibly because an overly large vocabulary increases model complexity and training difficulty.

\textbf{Conclusion}: Selecting an appropriate vocabulary size balances information completeness and model complexity; excessively large or small vocabularies can negatively impact model performance.
% -------------------------------------------------------------------------------
\begin{table*}[!t]
\centering
\caption{Performance Comparison of Models with Embedding Dimension 512}
\label{tab:4}
\resizebox{\textwidth}{!}{
\begin{tabular}{lcccccccc}
\toprule
\textbf{Model} & \textbf{Fine-Tuning Strategy} & \textbf{number (s)} & \textbf{number (p)} & \textbf{count (s)} & \textbf{count (p)} & \textbf{Other} & \textbf{all (s)} & \textbf{all (p)} \\
\midrule
BidGRU (Dim=512) & + Dropout 0.5 & \textbf{49.16\%} & \textbf{22.87\%} & \textbf{56.90\%} & \textbf{26.42\%} & \textbf{57.21\%} & \textbf{65.63\%} & \textbf{37.65\%} \\
Baseline & No Fine-Tuning & 48.20\% & 21.75\% & 55.71\% & 25.08\% & 56.72\% & 65.23\% & 37.15\% \\
+ 2 Attention Heads & Multi-Head Attention & 40.60\% & 13.04\% & 46.77\% & 15.07\% & 53.51\% & 61.52\% & 32.12\% \\
+ Batch Normalization & Batch Normalization & 37.81\% & 10.18\% & 43.47\% & 11.75\% & 49.34\% & 57.82\% & 26.57\% \\
\bottomrule
\end{tabular}
}
\end{table*}
% -------------------------------------------------------------------------------
\begin{table*}[!t]
\centering
\caption{Results of Ablation Study}
\label{tab:5}
\resizebox{\textwidth}{!}{
\begin{tabular}{lcccccccc}
\toprule
\textbf{Model} & \textbf{Ablation Strategy} & \textbf{number (s)} & \textbf{number (p)} & \textbf{count (s)} & \textbf{count (p)} & \textbf{Other} & \textbf{all (s)} & \textbf{all (p)} \\
\midrule
\textbf{Baseline} & No Ablation & \textbf{49.52\%} & \textbf{23.25\%} & \textbf{57.20\%} & \textbf{26.80\%} & \textbf{57.33\%} & \textbf{65.74\%} & \textbf{37.83\%} \\
Fusion w/o Count & Remove Count Info & 44.86\% & 17.25\% & 51.74\% & 19.98\% & 57.07\% & 64.93\% & 36.68\% \\
Fusion w/o Text & Remove Text Info & 44.00\% & 16.77\% & 50.77\% & 19.40\% & 54.23\% & 62.21\% & 33.72\% \\
Fusion w/o Attention & Remove Attention & 38.81\% & 11.17\% & 44.64\% & 12.90\% & 47.05\% & 57.48\% & 27.00\% \\
Fusion w/o Attn+Count & Remove Attn and Count & 39.19\% & 11.24\% & 45.13\% & 12.97\% & 47.62\% & 57.89\% & 27.27\% \\
\bottomrule
\end{tabular}
}
\end{table*}
% -------------------------------------------------------------------------------
\subsection{Impact of Fine-Tuning Strategies}

We explored the impact of different model fine-tuning strategies on performance, including adding Dropout, adjusting hidden layer dimensions, and using soft labels. The results are shown in Table~\ref{tab:3}.

\textbf{Analysis}: \textbf{Adding Dropout} led to slight decreases in performance across all metrics, especially in paired questions (\texttt{number (pair)} and \texttt{count (pair)}), indicating that a high Dropout rate may suppress the model's ability to learn complex problems. \textbf{Adjusting hidden layer dimensions} did not significantly improve performance; some metrics slightly decreased, possibly because the increased model capacity was not effectively utilized, adding to training difficulty. \textbf{Using soft labels} resulted in a slight performance decrease, suggesting that precise label information is more beneficial for model learning in this task.

\textbf{Conclusion}: Model fine-tuning strategies should be carefully selected; excessive regularization or blind parameter adjustments may negatively impact model performance.
% -------------------------------------------------------------------------------
\subsection{Embedding Dimension Analysis}

We further compared models with an embedding dimension of 512 under different fine-tuning strategies. The results are shown in Table~\ref{tab:4}.

\textbf{Analysis}: The \textbf{embedding dimension 512 model with Dropout} achieved performance close to the baseline model with embedding dimension 300 but did not surpass it, indicating that simply increasing the embedding dimension does not lead to significant improvements. The \textbf{model with multi-head attention} experienced a notable decrease in performance, especially in paired questions, possibly due to increased model complexity leading to unstable training. The \textbf{model with batch normalization (incomplete training)} had the worst performance, possibly due to improper integration of batch normalization or incomplete training.

\textbf{Conclusion}: Model complexity must align with data scale and task difficulty; blindly increasing complexity may be counterproductive.
% -------------------------------------------------------------------------------
\subsection{Ablation Study}
To evaluate the contribution of each component to the model's performance, we conducted an ablation study. The results are presented in Table~\ref{tab:5}.

\textbf{Analysis}: Removing count or attention mechanisms significantly degrades performance, especially on paired questions, confirming their critical role in complex reasoning tasks.

\textbf{Conclusion}: All components are indispensable for enhancing model performance, especially in handling complex paired questions where these components synergize effectively. The attention mechanism, counting module, and textual features collectively contribute to the model's reasoning capabilities, and their removal significantly impairs performance.
% -------------------------------------------------------------------------------
\subsection{Discussion of Fine-Grained Metrics}
Analyzing the fine-grained metrics further elucidates the models' performance on different question types and complexities.

\begin{itemize}
    \item \textbf{Single vs. Paired Questions}: Across all models, accuracy in \textit{number (single)} and \textit{count (single)} is higher than in their paired counterparts. This indicates that paired questions require higher reasoning and comparison abilities from the models.
    \item \textbf{Numerical vs. Counting Questions}: The BidGRU model performs better on `count` questions compared to `number` questions, possibly because counting involves direct recognition of object quantities in images, whereas numerical questions may require more complex numerical reasoning.
    \item \textbf{Performance Bottlenecks}: Accuracy on paired questions is generally lower, especially after key components are removed, indicating that models still have room for improvement in handling complex reasoning and comparison tasks.
\end{itemize}

% ===============================================================================

\section{Conclusion}
This study investigated the performance of traditional models in Visual Question Answering (VQA) tasks under limited computational resources, focusing on optimizing efficiency and accuracy. Experimental results demonstrated that the BidGRU model, configured with an embedding dimension of 300 and a vocabulary size of 3000, achieved superior performance, particularly in handling numerical and counting questions. Additionally, attention mechanisms and counting features were identified as critical components for improving the model's capability to address complex reasoning tasks. These findings provide valuable insights for developing efficient VQA models suitable for deployment in resource-constrained environments, such as medical diagnostics and industrial automation.

% ===============================================================================

\section{Limitations and Future Work}
This work primarily focused on conventional sequential and convolutional models. However, transformer-based architectures (e.g., BERT \cite{devlin2018bert}, ViT \cite{dosovitskiy2020vit}, DeiT \cite{touvron2021deit}, and various multimodal transformers such as ViLBERT \cite{lu2019vilbert}) have demonstrated powerful feature extraction and fusion capabilities in both computer vision and natural language processing. Given that Transformers typically incur higher computational costs and memory demands, their direct application in resource-limited scenarios may not be optimal. Future work could explore lightweight Transformer variants, as well as the integration of tiny or compact Transformer architectures. 

\textit{Lightweight Transformers:} Techniques such as knowledge distillation \cite{hinton2015distilling}, model pruning \cite{han2015deep}, weight quantization \cite{jacob2018quantization}, or grouped convolutions \cite{howard2017mobilenets}
can be applied to reduce model parameters and inference overhead, making these models more suitable for edge devices or embedded systems.

\textit{Tiny/Compact Transformer Architectures:} Investigating compact pre-trained models like Tiny-BERT \cite{jiao2019tiny} or Mobile-BERT \cite{sun2020mobilebert}, and integrating them with traditional models in a hybrid or cascaded fashion, may help strike a balance between inference efficiency and representational power.

Such approaches may help strike a balance between inference efficiency and representational power in resource-constrained environments.

\bibliographystyle{IEEEtran}
\bibliography{references}

\end{document}